\documentclass[a4paper,conference]{IEEEtran}
\ifCLASSINFOpdf
\else
\fi
\usepackage{times}
\usepackage{soul}
\usepackage{url}
\usepackage[hidelinks]{hyperref} 
\usepackage[utf8]{inputenc}
\usepackage[small]{caption}
\usepackage{graphicx}
\usepackage{amsmath}
\usepackage{amsthm}
\usepackage{booktabs}
\usepackage{algorithm}
\usepackage{multicol}

\usepackage{graphicx}
\usepackage{amsmath}
\usepackage{amssymb}
\usepackage{booktabs}

\usepackage{algorithm}
\usepackage{algorithmicx}
\usepackage[noend]{algpseudocode}
\algnewcommand{\LineComment}[1]{\State \(//\) #1}
\usepackage{caption}

\usepackage{float}

\usepackage{subcaption}
\usepackage{booktabs}
\usepackage{longtable}
\usepackage{adjustbox}
\usepackage{amsmath}
\usepackage{amssymb}

\begin{document}
%
\title{A Probabilistic Model Of Interaction Dynamics for Dyadic Face-to-Face Settings}

\author{\IEEEauthorblockN{Renke Wang}
\IEEEauthorblockA{Department of Computer Science\\
Rochester Institute of Technology\\
Rochester, New York\\
Email: rw7494@rit.edu
}
\and
\IEEEauthorblockN{Timothy Zee}
\IEEEauthorblockA{Department of Computer Science\\
Rochester Institute of Technology\\
Rochester, New York\\
Email: tsz2759@rit.edu}
\and
\IEEEauthorblockN{Ifeoma Nwogu}
\IEEEauthorblockA{Department of Computer Science\\
University at Buffalo, SUNY\\
Buffalo, New York \\ 
Email: inwogu@buffalo.edu}}


%


\maketitle

\begin{abstract}
Natural conversations between humans often involve a large number of non-verbal nuanced expressions, displayed at key times throughout the conversation. Understanding and being able to model these complex interactions is essential for creating realistic human-agent communication, whether in the virtual or physical world. As social robots and intelligent avatars emerge in popularity and utility, being able to realistically model and generate these dynamic expressions throughout conversations is critical. 
We develop a probabilistic model to capture the interaction dynamics between pairs of participants in a face-to-face setting, allowing  for the encoding of synchronous expressions between the interlocutors. This interaction encoding is then used to influence the generation when predicting one agent's future dynamics, conditioned on the other's current dynamics. FLAME features are extracted from videos containing natural conversations between subjects to train our interaction model. We successfully assess the efficacy of our proposed model via quantitative metrics and qualitative metrics, and show that it successfully captures the dynamics of a pair of interacting dyads. We also test the model with a never-before-seen parent-infant dataset comprising of two different modes of communication between the dyads, and show that our model successfully delineates between the modes, based on their interacting dynamics.
\end{abstract}

%
\IEEEpeerreviewmaketitle

\section{Introduction and Motivation}
\label{sec:intro}
Understanding social networks, which involve people and their interactions, is at the core of advancing human social intelligence. While the computational analysis of digital social network data (via social media sites) has enjoyed significant progress in this era of big-data and deep-learning technologies, the same cannot necessarily be said for comprehensively analyzing face-to-face interactions within dyadic or small groups; the types that exist in everyday occurrences such as e.g. doctor-patient, domestic partners, mentor-mentee, counseling sessions, work teams, friends, etc. 

Although the computational analysis of digital social media  has proven invaluable in elucidating many general patterns, much of this type of data about people and their behaviors is written, hence devoid of a larger range of emotion, it is composed entirely of asynchronous communication and we only learn about the kinds of people who use social media, rather than about human behavior in general. On the other hand, face-to-face communication is a highly interactive, synchronous process where participants mutually exchange and interpret messages, in real-time \cite{lpm2011} through various channels including facial expressions. 

\begin{figure}[ht!]
	\centering
	    \vspace{2mm}
		\includegraphics[width=0.98\linewidth]{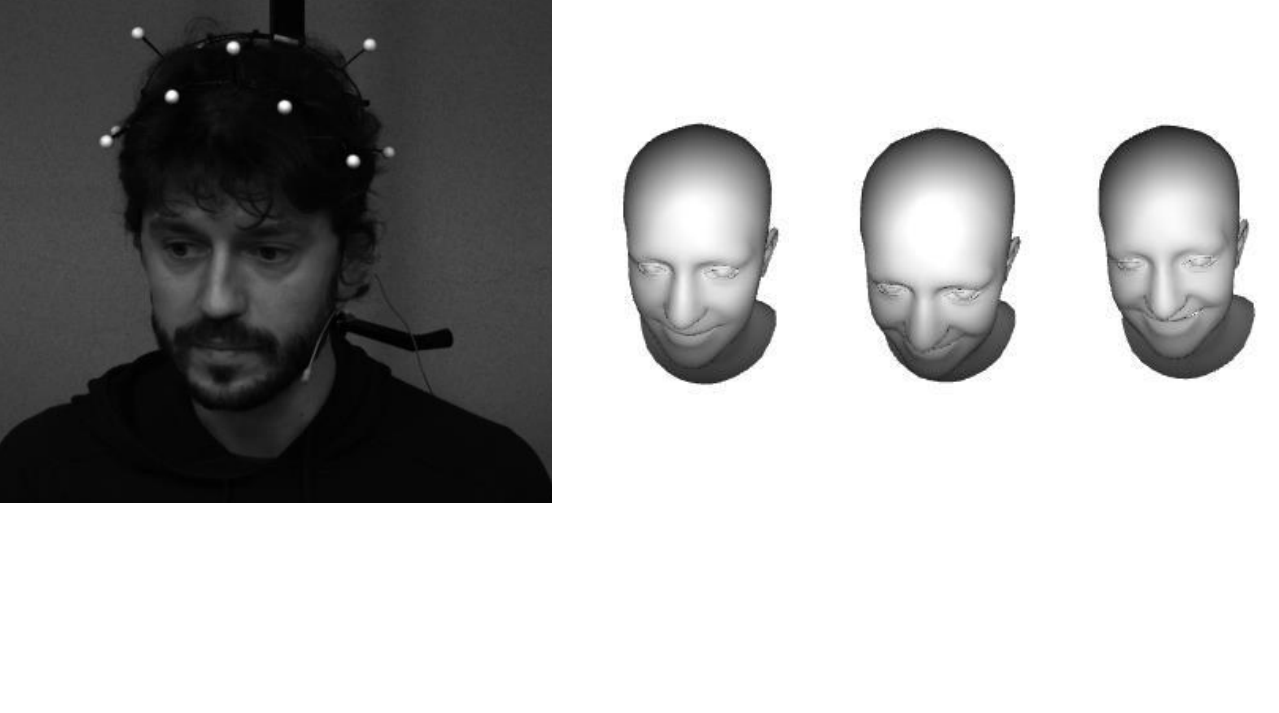}
		\vspace{-5mm}
		\caption{Left: Subject from MAHNOB dataset; right:  FLAME features on the subject's avatar, furnished with the corresponding facial expressions and head dynamics.}
		\label{fig:diffshape}
\end{figure}
According to Kendon \textit{et al.} \cite{f2f75}, face-to-face interaction is a basic element of the social system, forming a significant part of one's socialization and experience gaining. Even in this era of increased physical distancing, face-to-face interactions have not necessarily reduced, rather, they have moved from the physical to more virtual settings like video conferencing platforms \cite{vandenberg2021comparison}. 

In the social psychology literature, \emph{interactional  synchrony} is defined as \emph{the temporal coordination of micro-level social signals between two or more people communicating in a social setting}. For example, if one person smiles, in a synchronous interaction, the conversation partner would involuntarily smile in return. Or if one person nods continuously, the partner might find themselves nodding unkowlingly. Synchrony occurs at various nonverbal communication channel levels.

Synchrony in communication  plays an important role in maintaining positive social relationships among people since it indicates increased  affiliation,  rapport  and  feelings  of  empathy \cite{Bernieri91,chameleon99,chimps15}. 

While humans generally do well understanding these complex dynamics intuitively, it is not trivial to model or generate these actions on demand for use in human-agent communication. As remote agents are becoming more common, the demand for machine learning in this space can allow for improved and more natural communication between humans and agents such as virtual avatars, social robots, etc. 

\emph{To this end, the main contribution of this work is a generative model that encodes the dynamics between two conversational agents that are exhibiting interactional synchrony, and uses this encoding to influence the generation of the future dynamics of a conversational agent, when conditioned on its interlocutor agent.} \\

We provide metrics to show the efficacy of our model, and also perform an additional test that demonstrates that our model successfully delineates between interactions that involve synchrony and those that do not.

\section{Related Work}
\label{sec:relwork}
\subsection{Face Representation}
Although face representation via 2-D landmarks, using typically 68 key points, has been popular in the computational literature for face-based analysis, however, when dealing with a face with a wide angle viewpoint, some or many of these points may not be visible. To combat this, 3-D face representations have become more popular. 
3-D landmarking is still often too shallow and cannot capture a high fidelity map of the face, including elements like micro-expressions. 

To overcome this limitation, 3DDFA \cite{zhu2017face} fits a 2D face onto a 3D morphable model (3DMM) \cite{blanz1999morphable} to obtain a dense representation of a human face, and build a 3-D mesh based on the input image. While these advances are beneficial, extracted dense landmarks contain a large number of points, in the order of tens of thousands, and are challenging to use to represent frames in videos. 

FLAME \cite{li2017learning} was thus introduced to mitigate some of the drawbacks of 3DDDFA. FLAME uses different vectors to represent a human face, including the shape, rotation, pose, and expression vectors. All vectors are independent from each other, allowing for independently manipulating or removing unnecessary information from the extracted features. Another key advantage of FLAME is that it also provides a decoder, which can directly convert a set of FLAME parameters back into a 3-D mesh. For these reasons, for we opt to employ FLAME as the feature extraction technique we use to feed into our probabilistic interaction dynamic model.  

\subsection{Interaction video generation}
Previous work has focused on the generation of human gestures based on various conditions:
\cite{kucherenko2019analyzing} use an autoencoder to generate corresponding human motion based on the given speech audio. Other works like \cite{yoon2019robots} use a recurrent neural network to learn a mapping from text to body gesture. Other human gestures like generating human action sequences form labels are explored in \cite{petrovich2021action}.

\begin{figure*}[h!]
	\centering
	\vspace*{0mm}
		\includegraphics[width=0.9\linewidth]{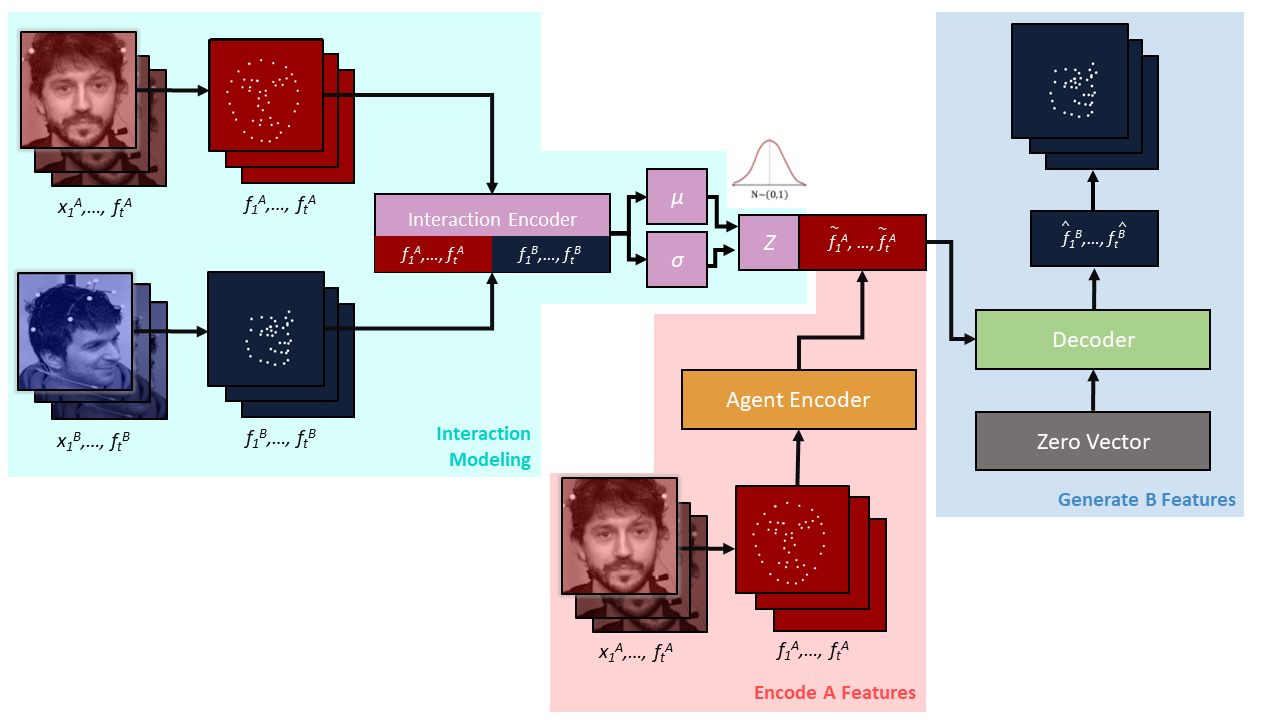}
		\caption{In our proposed probabilistic interaction dynamic model, we use an Agent Encoder $E^{A}$ to encode $F^{A}$ and use the output to represent Agent A's action $\widetilde{F}^{A}=[\widetilde{f}^{a}_{t}]_{t=1:T}$. We also use an Interaction Encoder $E^{I}$ to learn a representation of the interaction dynamics between Agent A and Agent B, and project the distribution of the interaction label into a Gaussian distribution. We concatenate Agent A's representation with interaction label $z$ and feed them into the Agent Decoder $D^{a}$ to generate $\hat{F}^{a}$.  }\label{TVAE}
\end{figure*}

While these works serve valuable roles in creating realistic human actions, we instead focus specifically on the generation of interactive dynamics, where another agent has a direct impact on how the agent continues to act over time. In this area, \cite{dermouche2019generative} build an interaction system, which could interact with humans in real-time. The extract expressions and audible speech as a condition to predict the agent's action. They create a model based on long short-term memory models and use conversational state (Current speaker at the given time step) as a label to help the model decide whether to take both participants' behavior into consideration. \cite{ahuja2019react} use two LSTM encoders to encode monadic and dyadic dynamics separately and use a Dyadic Residual Attention Model (DRAM) as a gate to decide whether the model should focus more on the agent motion or the interlocutor motion. Another work, \cite{feng2017learn2smile} firstly detect the landmark from both participants, and cluster on all the landmarks to obtain all possible motion, they then use a variational auto encoder to predict combined future movements. \cite{jonell2020let} encoded speech and face gesture sequence from both participants, with a normalizing flow, and predict the agent's future face gesture. \\
We create an end-to-end probabilistic model that contains three components (i) to model the dyadic interaction, (ii) to encode the dynamics of an input agent within the context of the learned dyadic interaction, and (iii) to decode and generate the interlocutor agent's response based on the encoded inputs. This results in the automated generation of synchronous dyadic face-to-face interactions. 
A visual depiction of the system is shown in Figure \ref{TVAE} and displays how interaction modeling is combined with an input agent's features to produce the  dynamics of the responding interlocutor agent.    


\section{Methodology}
\label{sec:method}

\subsection{Problem Definition}
Given a sequence of one agent’s (agent A) facial gestures $F^{A}=[f^{A}_{t}]_{t=1:T}$, we aim to generate the corresponding facial gesture sequence $\hat{F}^{B}=[\hat{f}^{B}_{t}]_{t=1:T}$ that the other agent (agent B) might exhibit in the interaction.

\subsection{Probabilistic Interaction Dynamic Model (PIDM)}
To successfully model complex dynamics we propose a three piece probabilistic model containing an interaction encoder, an agent encoder, and an agent decoder. The interaction encoder takes both agents' facial dynamics into account to establish a relationship between them. The agent encoder passes in encoded features in conjunction with the interaction encoder's sampled vector and is passed into the decoder to create corresponding dynamics for the other agent's next frame dynamics. The proposed model is shown in Figure \ref{TVAE}.

\subsubsection{Agent Encoder} 
The agent encoder $E^{A}$ receives agent A's FLAME sequence $F^{A}$ and outputs an embedded representation that the model will use with the embeddings for the interaction encode to decode for the alternative agent's dynamics for the same time sequence. $\widetilde{F}^{A}$ for agent A's action. 

\subsubsection{Interaction Encoder} 


The input of our interaction encoder $E^I$ is the concatenation of $F^{A}$ and ${F}^{B}$. At the beginning of the sequence, we add two special markers: [var] and [mean]. It is similar to the [cls] prepended marker introduced in the Bert \cite{devlin2018bert}, and ViT \cite{dosovitskiy2020image}, which learns a global representation of two sequences, and captures the pattern of interaction between the two sets of sequences. We use the output representation of these two markers as the variance and mean value, reparameterize with a random $Z$ vector, and use it as our interaction label.

We create an interaction label as interactions and responses among individuals may vary greatly for the same set of actions. We use $Z$ to capture that kind of difference. Once the reparameterization projects the interaction label into a Gaussian distribution, we randomly sample a $Z$ vector from the Gaussian distribution as our interaction label in training. 

\subsubsection{Agent Decoder} 
The input of our Agent Decoder $D^{A}$ will be a set of positional embedded zero vectors. Moreover, the cross attention layer will receive the concatenation of the $Z$ and $\widetilde{F}^{A}$. The output $\hat{F}^{B}$ is supposed to be as close as possible to ${F}^{B}$.

\subsection{Training}
The loss function of our PIDM model consists of the following two parts.

\subsubsection{Reconstruction Loss}
As we are using FLAME parameters in our model, we can easily define a reconstruction loss $L_{r e c}$ between $\hat{F}^{B}$ and ${F}^{B}$. We do an ablation study for both L1 and L2 loss to see which works better on the reconstruction loss, and the result shows the L1 loss gives us a better result. Further details are provided in the discussion section. The reconstruction loss $L_{r e c}$ can be formulated as follows:
\begin{equation}
\begin{aligned}
L_{r e c}=E_{F^{a}, \hat{F}^{a}}\left[\left\|F^{B}-\hat{F}^{B}\right\|_{1}\right]
\end{aligned}
\end{equation}
\subsubsection{KL-Loss}
Similar to other VAE models, we also have a KL-loss $L_{KL}$ in our loss function to ensure the interaction label has been projected into Gaussian distribution. We minimize the distribution between the Gaussian distribution and the distribution defined by our mean and variance from the interaction encoder.

Finally, the complete loss function will be:
\begin{equation}
L=L_{RC}+\lambda_{KL}*L_{KL}
\end{equation}
Here the $\lambda_{KL}$ is used to ensure the $L_{RC}$ and $L_{KL}$ are on the same scale.

For the reconstruction loss, we experiment with both L1 and L2 losses. We find empirically that L1 loss performs better and use that in our experiments. When training with l2 loss, the reconstruction loss fluctuates and is hard to converge. For an interaction, various people will have a different responses to the same set of actions. For example, when two participants are achieving agreement in the interaction, the participant might nod their head in response to the other's speech. However, there is no ground truth for when the head nod should happen. However, in our model, because we have the target motion for agent B, if the target motion nods its head on the 2nd second but generates a motion nod on the 3rd second, the L2 will give a big loss value. However, in common sense, both of the motions are correct. On the other side, L1 will give a relatively small loss because it is not too sensitive to the `wrong' example.

\begin{figure*}[t]
	\centering		\includegraphics[width=0.90\linewidth]{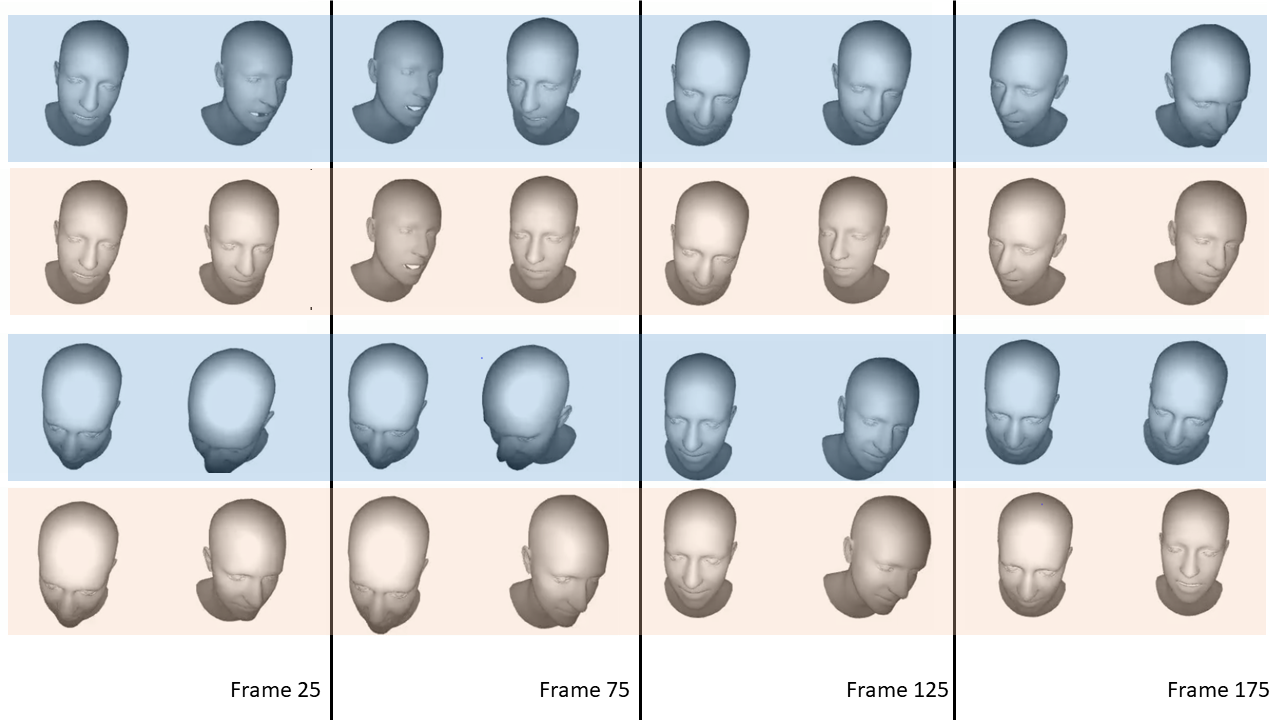}
		\caption{
		The first and third row (in blue) show the FLAME features of a pair of agents communicating, extracted from ground truth samples. The second and forth rows (in orange) show an input agent (from ground-truth) along with the generated interlocutor agent, that was created to response to the input agent's dynamics. For display purposes, the left avatar is the input and is the same for both ground-truth and the  generated sequences, while the right avatar is the reacting/responding agent.}\label{fig:results}
\end{figure*}
\subsection{Generating}
After training, we project the interaction condition label onto a Gaussian distribution. Thus, we can randomly sample a $Z$ from $N(0, 1)$ and let it be the interaction condition. Once we get $Z$, we can repeat what we did in the training process. That is to concatenate Z with $\hat{F}^{A}$ and feed it into Agent Decoder to generate the predicted response $\hat{F}^{B}$.

\subsection{Metrics}
To evaluate our model, we used two metrics. Each of the metrics focus on a different aspect of the model performance.
\subsubsection{Fréchet Video Distance}Fréchet Video Distance (FVD) \cite{unterthiner2018towards} can be used to mimic human judgment of video similarity. It measures the distance between the embedding of generated video and real video, which is used to train the generator. In order to calculate the FVD score, we use the FLAME decoder to decode agent A and agent B's FLAME sequence into videos and concatenate both videos side by side. Then use the concatenated video as compared samples. It should be noted that when we calculate the FVD score for ground truth, we divide the training set into two parts with no intersect. Then calculate the FVD value between these two parts, so the score will not be 0.

\subsubsection{Mean Absolute Error} Mean Absolute Error (MAE) is the other criterion we use. We first normalize the pose, expression, and rotation in the FLAME sequence to ensure they are of the same magnitude order. Then we calculate the MAE between the ground truth and the generated result for each feature and add them up as the final MAE value. This was done because 50 out of the 59 dimensions in the FLAME parameter are used to represent an expression. If we directly calculate MAE on the FLAME parameter, the expression will dominate the result. After calculating MAE for each feature independently, we believe that MAE will be a good representation of our model's ability to reconstruct the original FLAME sequence.

\section{Implementation Setup}
\label{sec:expts}
\subsection{Interaction Training Data}
In our work, we use the MAHNOB Mimicry Database \cite{bilakhia2015mahnob}. 
This dataset contains 53 pairs of face-to-face conversations. Each conversation is carried out by a different pair of participants and focuses on different topics. We select sections 1 and 2 as test sets, 3 and 4 as the validation set, and all other sessions become the training set. Each video contains approximately 10 to 20 minutes of interaction between the two agents. For training purposes, each video is cut into 10 second clips, where a sliding window with a step size of one is used to generate additional samples of data. When the model is training, we randomly designate one participant as the model input, which we call Agent A. The other participant who will respond to the first agent is called Agent B. Our proposed model aims to learn to predict how Agent B will react to Agent A in the course of a conversation between them.

\subsection{Feature Extraction}
In order to obtain the FLAME sequence, we use the FAN model \cite{bulat2017far} to detect the landmark from the original video. We then use the detected landmarks to create a bounding box to crop the face from the original video frame. Then the cropped face is fed into Ringnet \cite{sanyal2019learning} to obtain the final FLAME parameter. The expression vector in FLAME is obtained by Principal component analysis (PCA), so we choose only the first 50 parameters to represent the face. We also include the six-dimensional pose vector three-dimension rotation vector in our final face representation vector. Hence, the final face representation is a 59-dimensional vector. We did not include the shape vector of FLAME in our training data because the face shape usually has no effect during the conversation. 

After extracting FLAME from the video, We did a 5 step running average on the FLAME sequence to smooth the resulting video, since FLAME features are obtained frame by frame, which might introduce unexpected noise. After smoothing, we cut each pair of FLAME sequences into 10-second sub-sequences and to obtain our training data.


%

We used the FLAME decoder provided by \cite{li2017learning}, which could directly generate target 3D mesh from FLAME parameters. Also, because of the characteristic of FLAME, we can manipulate the shape of the face without affecting the generated movement. Figure \ref{fig:diffshape} (right) shows three different randomly selected face shapes for the agent.

\subsection{Model Configuration and Training}
Our interaction encoder $E^I$, agent encoder $E^A$, agent decoder $D^A$ are using Transformer as the backbone, and they share many hyper-parameters: 6 layers of Transformer blocks; the dropout rate in each Transformer block is 0.1; the learning rate is 1e-5 without decay; the batch size is 16; the $\lambda_{KL}$ equals 0.001. All of the parameters are optimized with the Adam optimizer and trained for 800,000 iterations.

\section{Experiments and Results}
\subsection{Baseline Experimental Setup}
The closest to this work is the model by \cite{jonell2020let}, but they focus on integrating speech generation where the avatar's face, specifically the mouth, is trained to adjust according to the words being spoken. It is not trivial to tease apart the encoded speech formation in this model, hence comparing with it is non-trivial. 

For this reason, as we could not find other works involved in the same task, we manually created some baselines to evaluate with our model against. Here are the baselines we selected:
We generate methods to test the model to serve as baselines to use for comparison to determine if we are capturing real dynamics portrayed between pairs of two dyadic agents. We create the following methods to serve as baselines: 

\noindent \textit{Noise}: Randomly sample from Gaussian noise to use as the prediction for Agent B's response. \\
\textit{Mirror}: Copies the identical FLAME sequence of agent A with 3 second delay to use as Agent B's response. \\
\textit{Random}: Chooses a random equally likely FLAME sequence from the training set as Agent B's response. \\
\textit{Ground truth}: Agent B's actual response from the dataset.\\

Except for the ``Noise'', all other baselines are real sequences from the dataset. The only difference is that the simulated agent B in "Mirror" and "Random" do not really interact with agent A.
\begin{table}[htbp]
\footnotesize
  \centering
\begin{tabular}{llrr}
      &       &  \\
\toprule
\textit{Metrics} & \textit{Description} & \textit{FVD$\big\downarrow$.}  & \textit{MAE$\big\downarrow$.} \\
\midrule
Noise & $x \sim \mathcal{N}(0,1)$ & 1229.10 &  3.28\\
\midrule
Mirror & 3s delay & 676.58 &  3.11\\
\midrule
Random & Any frame $\in X$ &572.26 &  3.11\\
\midrule
Ground truth & - & 342.02 & 0 \\
\toprule
PIDM (Proposed) & - & 542.62 &  2.60\\
\bottomrule
\end{tabular}%
  \caption{Quantitative results for baselines and our model.}
  \label{tab:FVD&MAE}%
\end{table}%

Comparing with other groups in the baseline study, Table \ref{tab:FVD&MAE} shows that PIDM successfully models the interaction between two people. It is able to generate correct (low MAE) and vivid (low FVD) responses, to the given input facial gestures.  

Figure \ref{fig:results} shows 2 examples of ground-truth dynamics along with the PIDM outputs, comprising of the generated responses of an interlocutor agent when one of the ground-truth agents is fed to the model as input. The real and simulated videos are included in the Supplementary material.

\subsection{Still-Face Analysis}
In Pediatric Psychology, the still-face experiment studies the reaction of face-to-face dynamics between a parent and infant under varying experimental conditions. Such experiments are useful in monitoring the mental health of infants in the early stages of life \cite{beebe2005mother}. 

\begin{figure}[h]
	\centering
		\includegraphics[width=0.85\linewidth]{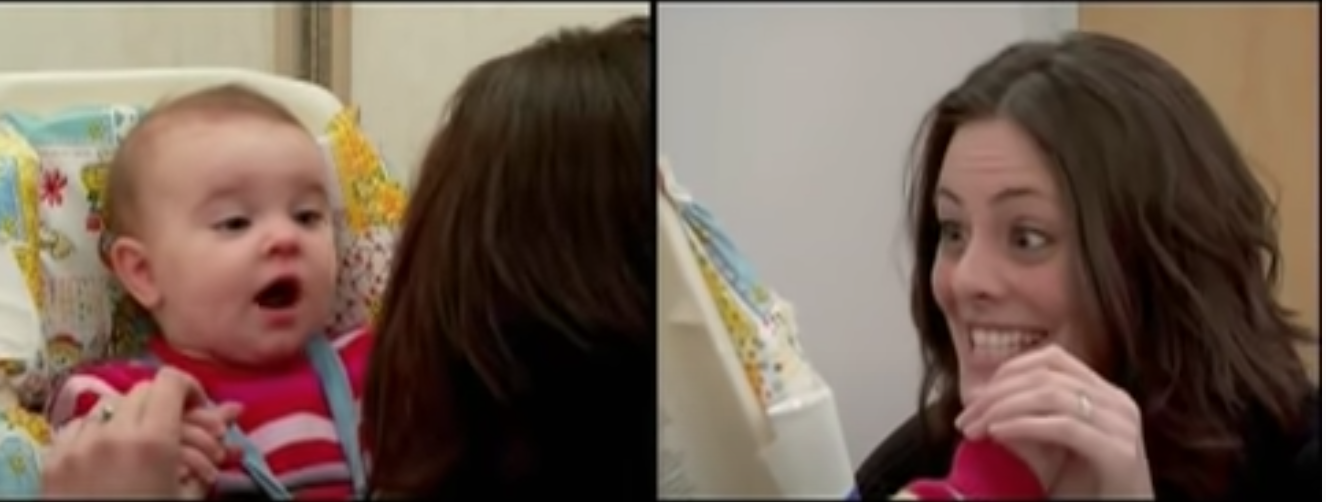} \vspace*{2mm}
        \includegraphics[width=0.85\linewidth]{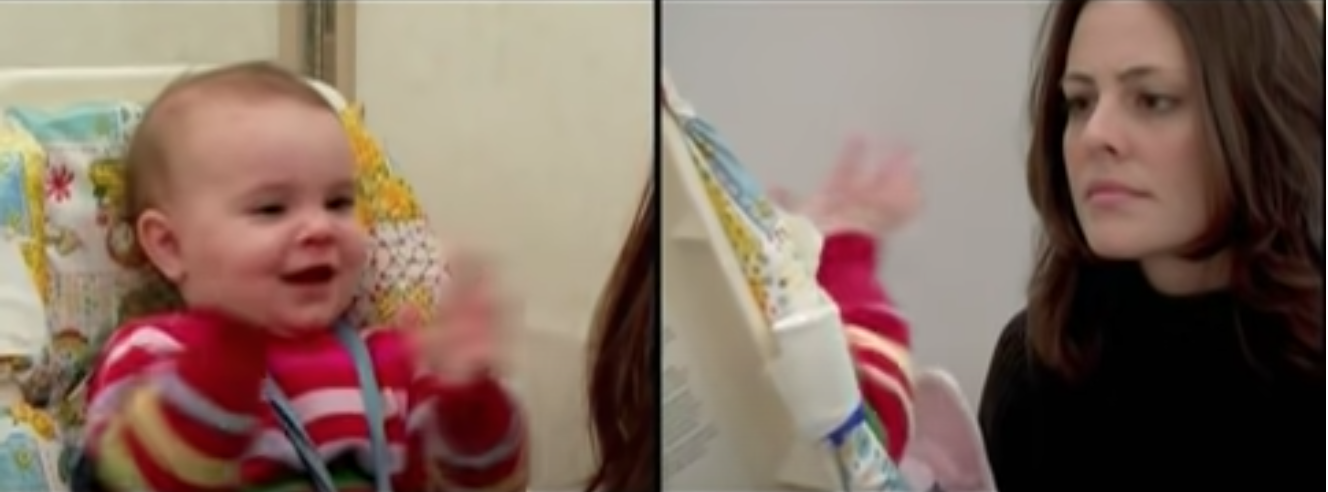}
		\caption{Still-face experiment between a mother and baby. The top images (the face-to-face condition) shows the typical dynamics between the mother and baby exhibiting interactional synchrony; while the bottom images (the still-face condition) shows the atypical, non-synchronous dynamics between the same mother and baby, where the mother keeps a still and stoic expression, regardless of her baby's dynamics.
		\tiny{(Image obtained from \url{https://www.youtube.com/watch?v=apzXGEbZht0})}}\label{fig:face2face}
\end{figure} 

In the first experimental condition, the parent  interacts with the infant exchanging positive emotional expressions in a way the infant is familiar with. This is called the face-to-face condition. In the next condition, the parent is instructed to discontinue reacting emotionally to the infant, irrespective of any dynamics the infant displays. This is called the still-face condition. The top image in Figure \ref{fig:face2face} shows an example of the interactional synchrony in the face-to-face condition, while the bottom image, still-face condition, shows the absence of synchrony in this interaction, as the mother maintains a stoic facial expression. The face-to-face and still-face conditions are captured as separate video recordings for this work.


We apply our model which is trained on data which has synchrony dynamics, and test on both the face-to-face and still-face datasets. We study the resulting model errors and this provides us with another way to evaluate the model. 
We find that when comparing test errors between the two types of interactions, the face-to-face data yields significantly less error than the still face data, indicating that our model expects synchronous reactions between the two agents, and not one where an agent holds a still face while the other is emoting. 


\begin{figure}
     \centering
     \begin{tabular}{cc}
     \includegraphics[width=.48\linewidth]{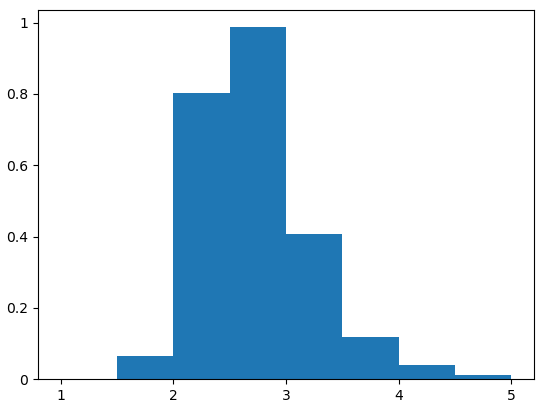} \hspace*{-5mm}&
     \includegraphics[width=.48\linewidth]{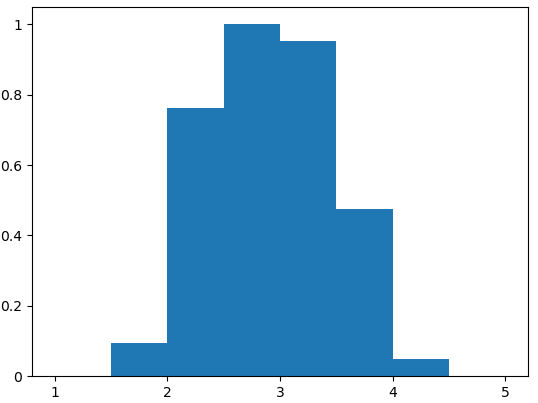}\\
    \end{tabular}
    \caption{Comparing the errors from using face-to-face data (left) versus the still face data (right) when running inference with the trained interaction model. The y-axis is the normalized frequency between 0 and 1 and the x-axis is the error range. The area under the curve is the useful information.}\label{fig:errors}
\end{figure}

\subsection{Human Judgement Test}
For this test, we showed our test participants two videos each having a pair of avatars furnished with FLAME features. The first video consisted of just the FLAME features from the actual interacting dyads, while the second video consisted of similar FLAME videos, except that the facial expressions and head dynamics of one of the dyads was generated in response to the partner. Twenty random video pairs (generated and ground-truth) were shown.

\begin{figure}[h]
	\centering
		\includegraphics[width=0.85\linewidth]{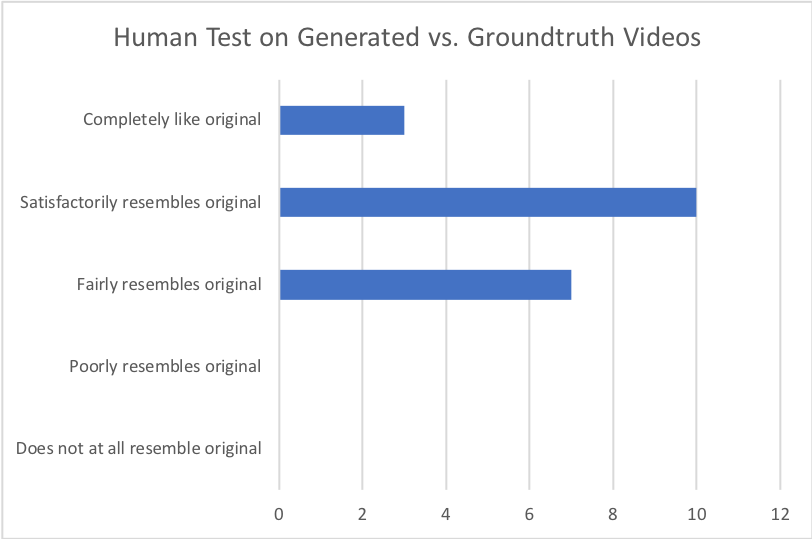} 
		\caption{Results of comparing generated interaction videos with ground truth ones}\label{fig:humantest}
\end{figure}

The test participants rated the two videos to highlight how similar their behavior patterns were. This was based on appropriate facial expressions and dynamics, and how much they perceived that the generated model was interacting and responding to the interlocutor. The rating scales are shown as y-axis labels in Figure \ref{fig:humantest}, the results of the human experiment. 

\section{Discussion and Conclusion}
\label{sec:discussion_conclusion}


In this paper, we introduce a model which successfully generates human-like response based on the visual cues of a given human interlocutor. We demonstrate that FVD can be used to detect the presence or absence of interactions when presented with a pair of signals, thus providing a set of quantitative metrics for similar tasks.


In our analysis, we used FVD, a derivative of the Fréchet inception distance (FID) \cite{heusel2017gans} for evaluation. Both FVD and FID use the same technique to measure the similarity between two samples groups. They feed both the generated and the real samples into a pre-trained GoogLeNet  \cite{szegedy2015going} model, and calculate the similarity between the intermediate layers' output. Since different layers in the neural network provide a different level of abstraction for the input samples, it is generally believed that the results are representative of how similar the generated samples are to the original samples \cite{lucic2018gans}. 
FVD was shown to sufficiently mimic human judgment of Mirror, Random, Ground-truth and Generated videos, thus providing a quantitative metric to better understand whether the generated videos are naturalistic and interactive.

The still-face experiment further demonstrated that our proposed model learned how an agent responds synchronously to an interacting input agent. As shown in Figure \ref{fig:errors}, the face-to-face synchronous condition yielded significantly less errors than the still-face, no-synchrony condition. This shows that the model is tuned to perform better in the presence of synchrony. 

Lastly, from further analysis, the results of a human judgement test were quite high, showing that the test participants did not find any of the generated video pairs to be significantly different in behavior pattern from the originals. As much as \textbf{65\%} were judged to have behavior patterns that satisfactorily resembled original or better, and \textbf{100\%} were found to be fairly similar to the original or better. 

For our future work, we will aim to improve the interaction embedding $Z$ from the latent space, to generate even more realistic interaction responses to a given input agent. 

\small
\bibliographystyle{IEEEtranS.bst}
\bibliography{refs.bib, refs1.bib}

\end{document}